\documentclass[runningheads]{llncs}
\newcommand{\mymodel}{CHIS}
\usepackage[T1]{fontenc}
\usepackage[scr=rsfso]{mathalpha}
\usepackage{graphicx,verbatim}
\usepackage{amssymb}
\usepackage{circledsteps}
\usepackage{amsmath}
\usepackage{booktabs}
\usepackage{multirow}
\usepackage{xcolor}
\usepackage{threeparttable}
\usepackage{hyperref}

\usepackage{arydshln}

\begin{document}
\title{Controllable Histopathology Image Synthesis with Training-free Structural Initialization and Textural Modulation}
%

\author{
Yuheng Qiu\inst{1}
\and Jingyi Luo\inst{1}
\and Chenfei Ye\inst{1}
\and Ting Ma\inst{1}
\and Jianfeng Cao\inst{1}\thanks{Corresponding author: caojianfeng@hit.edu.cn}
}

\authorrunning{Qiu et al.}

\institute{
School of Biomedical Engineering, Harbin Institute of Technology (Shenzhen), Shenzhen, China\\
\email{\{qiuyuheng, 25b963007\}@stu.hit.edu.cn}\\
\email{\{yechenfei, tma, caojianfeng\}@hit.edu.cn}
}

\maketitle              
\begin{abstract}
Deep learning has demonstrated remarkable success in high-throughput histopathology image analysis. However, the performance of learning-based models critically depends on the quality and size of annotation by expert pathologists—a resource-intensive and time-consuming process. To address the limitations of data scarcity and annotation burden, several methods have been proposed to synthesize paired histopathology data. Nevertheless, these frameworks typically still require annotation data, albeit in reduced quantities, to impose structural constraints during training. In this work, we present \mymodel, a plug-in framework that guides the sampling trajectory of a pretrained diffusion model through two key stages: structural initialization at the start and textural modulation during generation. The initial noise state is refined by fusing the phase information from a prior mask with the amplitude of Gaussian noise in the frequency domain, yielding a structurally-informed starting point. During the reverse diffusion process, we adaptively modulate both coarse- and fine-grained textures at different wavelet decomposition levels. This enables a diffusion model pretrained solely on unlabeled images to generate outputs that align with prior structural masks while preserving the reference tissue style. We conducted extensive experiments demonstrating the superiority of \mymodel\ in generation fidelity and its substantial benefits for downstream segmentation tasks. Code is available at https://github.com/IBIL-Code/CHIS.

\keywords{Histopathology Analysis  \and Guided Image Generation \and 
 Diffusion Model \and Unsupervised Learning}

\end{abstract}
\section{Introduction}
Histopathology remains the gold standard for diagnosis and grading because of its ability to distinguish benign from malignant tissue \cite{song2023artificial}. Learning-based algorithms now enable high-throughput analysis of whole-slide images, supporting automatic annotation and AI-assisted reporting with user-defined quantitative metrics that reduce pathologists' workload \cite{ding2025multimodal,janowczyk2016deep}. However, model accuracy depends critically on the quantity and quality of labeled training data, which is expensive and time-consuming to obtain in routine clinical workflows. Synthetic data generation therefore offers a practical solution to alleviate annotation scarcity. While generative adversarial networks (GANs) can learn with weak or unpaired supervision, they often introduce artifacts that compromise diagnostic realism \cite{butte2022sharp,tschuchnig2020generative}. Diffusion models typically achieve higher fidelity and better conditional control, but most existing approaches require paired supervision for training \cite{sdm,Nudiff}. Cycle-diffusion methods attempt unpaired translation but unfortunately suffer from noisy-clean domain mismatch, substantially increasing computational cost \cite{zhu2025cycle,zou2025cyclediff}. Furthermore, standard Gaussian initialization inevitably induces structural misalignment during generation \cite{bhosale2025pathdiff,li2023zero}.

\begin{figure}[t]
    \centering
    \includegraphics[width=\linewidth]{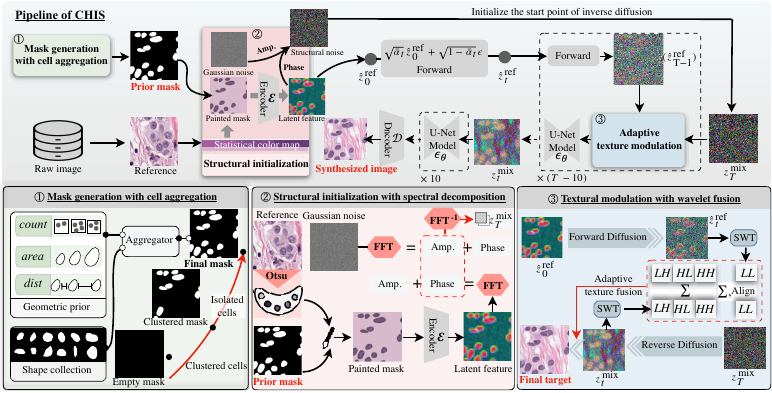}
    \caption{Overview of our proposed \mymodel\ for controllable histopathology image synthesis.}
    \label{Fig:pipeline}
\end{figure}
In this work, we propose \mymodel, a training-free framework that guides pretrained diffusion models to generate structure-aligned histopathology images (Fig. \ref{Fig:pipeline}). Given a diffusion model trained solely on unlabeled histopathology images, our \mymodel\ first generates a prior mask to regulate the spatial location of generated cells. For the starting point of the reverse sampling process, we leverage the insight that phase information primarily encodes structural contours and refine the initialization by fusing the phase of the prior mask with the magnitude of Gaussian noise in the frequency domain. Subsequently, fine-grained textures progressively interact with the dynamic sampling states through adaptive aggregation of wavelet components. The enhanced diffusion model ultimately delivers synthesized data that are not only stylistically consistent but also structurally controllable. Comprehensive evaluations demonstrate the superiority of \mymodel\ in image quality, mask-structure fidelity, and downstream segmentation performance.

\section{Method}
In this section, we first introduce the pretrained diffusion model that serves as the backbone of \mymodel. We then present an efficient framework for generating prior masks. In the following two subsections, we sequentially describe our strategies for refining the initialization state and guiding the reverse diffusion sampling process.

\subsection{Pretrained Diffusion Model with Histopathology Images}
We select the Latent Diffusion Model (LDM)~\cite{ldm} as our generative backbone, which shifts the learning of complex image distributions from the high-dimensional pixel space $\mathbf{X}$ to a compressed latent space $\mathbf{Z}$. Given a histopathology image $x \in \mathbf{X}$, a pretrained VAE encoder $\mathcal{E}$ maps it to a latent representation $z_0 = \mathcal{E}(x)$. The forward diffusion process progressively adds Gaussian noise to $z_0$ over $T$ timesteps. For any timestep $t \in [0, T]$, the noisy latent $z_t$ is defined as:
\begin{equation}
z_t = \sqrt{\bar{\alpha}_t} z_0 + \sqrt{1-\bar{\alpha}_t} \epsilon, \quad \epsilon \sim \mathcal{N}(0, \mathbf{I})
\label{eq:forward}
\end{equation}
where $\bar{\alpha}_t$ follows a predefined noise schedule. The reverse denoising process follows the DDIM~\cite{ddim} formulation:
\begin{equation}
z_{t-1} = \sqrt{\bar{\alpha}_{t-1}}
\left(\frac{z_t - \sqrt{1-\bar{\alpha}_t}\,\epsilon_\theta(z_t,t)}{\sqrt{\bar{\alpha}_t}}\right)
+ \sqrt{1-\bar{\alpha}_{t-1}}\,\epsilon_\theta(z_t,t)
\label{eq:reverse}
\end{equation}
where the denoising network $\epsilon_\theta(z_t, t)$ predicts the noise added to $z_t$, allowing deterministic reconstruction of $\hat{z}_0$ from
$z_T \sim \mathcal{N}(0, \mathbf{I})$ step-by-step. 
The learnable model $\epsilon_\theta$ is pretrained solely on unlabeled histopathology images and remains frozen in our proposed \mymodel. Finally, the VAE decoder $\mathcal{D}$ produces synthesized images by $\hat{x}_0 = \mathcal{D}(\hat{z}_0)$. In the following sections, we will describe how \mymodel\ ensures the spatial distribution of cells in $\hat{x}_0$, starting from a random state $z_T$, can be precisely controlled by masks $y\in\mathbf{Y}$ generated in Section \ref{Sec:cellular_masks}.

\subsection{Generation of Histopathology Mask with Cell Aggregation}
\label{Sec:cellular_masks}
To facilitate the generation of histopathology masks, we propose an efficient framework for aggregating single-cell masks with geometric priors (\Circled{1} in Fig. \ref{Fig:pipeline}). Given a collection of cell masks, the synthesis of a histopathology mask $y$ can be formulated as an aggregation process constrained by three category-specific metrics: cell count, mask area, and inter-cell distance, denoted as $\{count, area, dist\}$. These metrics can be either derived from the mask collection or specified directly by pathologists.

To aggregate cell masks into a histopathology mask, we partition $count$ cells into two groups: clustered cells and isolated cells. We first generate cell clusters by sequentially sampling cells from the collection and positioning them together within the specified $dist$ threshold. Each cell cluster is then treated as a super-cell and combined with isolated cells from the collection. The distances between all cells and super-cells are maintained above the preset $dist$ threshold. Additionally, the size of each cell fluctuates around the geometric prior $area$. This generated mask benefits downstream segmentation models through densely distributed clustered cells while preserving mask fidelity through the inclusion of isolated cells.

\subsection{Structural Initialization with Spectral Decomposition}
During the reverse sampling process, the initial state $z_T$ largely determines the content of the denoised results. Rather than randomly sampling $z_T$, we incorporate structural features from the mask $y$—specifically, the contours of cell masks—into $z_T$ by decomposing $y$ using the Fast Fourier Transform (FFT). As described in \cite{zeng2025neuralremaster}, structural information is primarily preserved in the phase component. We therefore refine $z_T$ through structure-aligned phase fusion between the mask $y$ and Gaussian noise $z_T$ (\Circled{2} in Fig. \ref{Fig:pipeline}).

For the latent diffusion model, the encoder $\mathcal{E}$ transforms the input into a latent feature map. Although $\mathcal{E}$ can directly accept the mask $y$, we observe a significant distribution shift in $\mathcal{E}(y)$ when a binary mask $y$ differs substantially from the pretraining data (i.e., histopathology images). To address this issue, we apply an extract-and-fill procedure to roughly colorize the regions of $y$. Specifically, we first select a reference histopathology image $x^\text{ref}$ and apply Otsu thresholding to obtain its coarse segmentation in the optical density domain. The background regions of $y$ are then filled with the average color of the corresponding regions in $x^\text{ref}$. To enhance the textural features of foreground cells in $y$, pixels are colored according to the average color of pixels at the same boundary distance in $x^\text{ref}$. Finally, we obtain $\hat{y}$ with a color style similar to that of $x^\text{ref}$.

Based on the refined input $\hat{y}$, we obtain the latent feature $\hat{z}^\text{ref}_0=\mathcal{E}(\hat{y})$. The FFT operation $\mathcal{F}$ is applied to $\hat{z}^\text{ref}_0$ and $z_T$:
\begin{equation}
    \mathcal{F}(\hat{z}^\text{ref}_0)=A^\text{ref}\exp(j\phi^\text{ref}),\quad\mathcal{F}(z_T)=A^\text{noise}\exp(j\phi^\text{noise})
\end{equation}
where $A^{\cdot}$ and $\phi^{\cdot}$ denote the amplitude and phase parts, respectively. These two parts are subsequently mixed with
\begin{equation}
    A^\text{mix}=A^\text{noise},\quad\phi^\text{mix}=\Theta(r)\odot\phi^\text{ref}
    +(1-\Theta(r))\odot\phi^\text{noise}
\end{equation}
where $\Theta(r)$ is a low-pass filter that keeps phase only within a cutoff radius $r$. The structure-aligned state comes with inverse FFT $z^\text{mix}_T=\mathcal{F}^{-1}(A^\text{mix}\exp(j\phi^\text{mix}))$. With different $r$, we can control the degree of structural preservation regarding mask $y$. Additionally, the style of $z^\text{mix}_T$
can be easily changed by selecting another $x^\text{ref}$ even without any annotations.

\subsection{Adaptive Textural Modulation with Wavelet Fusion}
Beyond structural alignment, textural similarity also contributes to the fidelity of generated images and further reinforces structural consistency. Here we propose wavelet-based guidance to progressively enhance textural details in $z^\text{mix}_t$ and improve structure–texture coherence during the reverse diffusion process (\Circled{3} in Fig. \ref{Fig:pipeline}). We employ the stationary wavelet transform (SWT) $\mathcal{S}$ to decompose both the evolving latent $z^\text{mix}_t$ and the prior guidance $\hat{z}^\text{ref}_t$ into coarse- and fine-grained texture components, yielding
\begin{equation}
    \mathcal{S}(z^\text{mix}_t)=\{L^\text{mix}_t, H^\text{mix}_t\},\quad \mathcal{S}(\hat{z}^\text{ref}_t)=\{L^\text{ref}_t, H^\text{ref}_t\}
\end{equation}
where the coarse texture $L^\cdot_t=[LL^\cdot_t]$ and fine-grained texture $H^\cdot_t=[LH^\cdot_t\mid HL^\cdot_t\mid HH^\cdot_t]$ correspond to low- and high-frequency components in the frequency domain, respectively \cite{huang1998predictive}. 

Since the coarse texture $L^\text{mix}_t$ depicts the overall contours in $z^\text{mix}_t$, we aim to align it with $L^\text{ref}_t$. Hence, the distribution of $L^\text{mix}_t$ is shifted using the mean $\mu(\cdot)$ and standard deviation $\sigma(\cdot)$ via:
\begin{equation}
    \tilde{L}^\text{mix}_t = \sigma(L^\text{mix}_t)\,\frac{L^\text{ref}_t - \mu(L^\text{ref}_t)}{\sigma(L^\text{ref}_t)} + \mu(L^\text{mix}_t),
    \label{eq:adain}
\end{equation}
Then we modulate foreground and background textures based on three rules: i) Relaxed constraints on background textures; ii) Primary reliance on $\hat{z}^\text{ref}_t$ for low-frequency contour textures; iii) Reduced guidance on detailed high-frequency textures. Three hyperparameters $\gamma$, $\chi$ and $\zeta$ were set accordingly to control the degree of texture modulation. Thereafter, the final weights for modulating low- and high-frequency components are given by $\omega_L=\delta_{\{y,1\}}\chi+ \delta_{\{y,0\}}\gamma$ and $\omega_H = \delta_{\{y,1\}}\zeta+\delta_{\{y,0\}}\gamma$, where $\delta$ represents the Kronecker delta function. These two weights guide the modulation with weighted summation
\begin{equation}
    \tilde{L}^\text{mix}_t = \omega_L \cdot\hat{L}^\text{mix}_t + (1-\omega_L) \cdot L^\text{mix}_t, \quad
    \tilde{H}^\text{mix}_t = \omega_H \cdot H^\text{ref}_t  + (1-\omega_H) \cdot H^\text{mix}_t,
    \label{eq:blend}
\end{equation}
Finally, the inverse SWT transforms the components $\tilde{L}^\text{mix}_t$ and $\tilde{H}^\text{mix}_t$ back into the spatial domain $z^\text{mix}_t$, status for the next sampling step. This process repeats during the early sampling stage, until the textures are well consolidated by $\hat{z}^\text{ref}_t$.

\section{Experiments}
\subsection{Implementation Details}
While \mymodel\ is framework-agnostic across diffusion-based models, we
instantiate it on PixCell~\cite{pixcell}, a backbone pretrained on large-scale
histopathology images that synthesizes realistic tissue but leaves cellular
structure uncontrolled. We keep PixCell \emph{frozen}, preserving all parameters
and its foundation-feature conditioning; \mymodel\ acts only on the
sampling trajectory: it replaces the initial noise at $t{=}T$ with a phase-fused
latent, applies textural guidance over the first 40 steps
with $\gamma=0.2$, $\chi=0.95$, and $\zeta=0.6$, and lets PixCell run freely in
the final 10 steps.

We evaluate \mymodel\ on three histopathology datasets: MoNuSAC~\cite{monusac}, Kumar~\cite{kumar}, and PanNuke~\cite{pannuke}, with an 80/20 split for training/test data. For comparison, we select various baseline methods representing different approaches, including fully supervised models (NuDiff~\cite{Nudiff}, SDM~\cite{sdm}), GAN-based models (SynDiff~\cite{syndiff}, CycleDiff~\cite{zou2025cyclediff}), and training-free models (UGDM~\cite{ugdm}, ADMMDiff~\cite{admmdiff}). For all methods, the number of synthesized images matches that of the training split in each dataset. The synthesized images are assessed from three perspectives: (i) evaluate their alignment to prior masks using a binary segmentation model; (ii) investigate their benefits to downstream instance segmentation model; (iii) measure their fidelity compared to real datasets. Detailed experimental settings are introduced in the corresponding sections.
\begin{table}[t]
\centering
\caption{\textbf{Evaluation on the consistency between prior mask and synthesized image.}
(Best result in bold, second-best  underlined. Fully supervised models are listed for reference and not for ranking.)}
\label{tab:fs1_hd95}
{\footnotesize
\begin{tabular}{cc cc cc cc cc}
\toprule[1.5pt]
\multirow{2}{*}{Method} & \multirow{2}{*}{Training} & \multirow{2}{*}{\shortstack{Paired\\data}}
& \multicolumn{2}{c}{PanNuke}
& \multicolumn{2}{c}{MoNuSAC}
& \multicolumn{2}{c}{Kumar} \\
\cmidrule(lr){4-5} \cmidrule(lr){6-7} \cmidrule(lr){8-9}
& &
& FS1$\uparrow$ & HD95$\downarrow$
& FS1$\uparrow$ & HD95$\downarrow$
& FS1$\uparrow$ & HD95$\downarrow$ \\
\midrule[1pt]


\textcolor{gray}{SDM~\cite{sdm}} & \textcolor{gray}{$\circ$} & \textcolor{gray}{$\circ$}
& \textcolor{gray}{0.7462} & \textcolor{gray}{30.8448}
& \textcolor{gray}{0.8167} & \textcolor{gray}{24.9139}
& \textcolor{gray}{0.7669} & \textcolor{gray}{19.4987} \\

\textcolor{gray}{NuDiff~\cite{Nudiff}} & \textcolor{gray}{$\circ$} & \textcolor{gray}{$\circ$}
& \textcolor{gray}{\textbf{0.8232}} & \textcolor{gray}{\textbf{22.3510}}
& \textcolor{gray}{0.8264} & \textcolor{gray}{22.7506}
& \textcolor{gray}{\textbf{0.7953}} & \textcolor{gray}{\textbf{17.1287}} \\

\addlinespace[3pt] 
\hdashline
\addlinespace[3pt] 

SynDiff~\cite{syndiff} & $\circ$ & $-$
& 0.0586 & 90.0501
& 0.0025 & 243.4370
& 0.2150 & 37.6211 \\

CycleDiff~\cite{zou2025cyclediff} & $\circ$ & $-$
& 0.0874 & 62.3175
& 0.2002 & 80.4365
& 0.1343 & 45.1336 \\

UGDM~\cite{ugdm} & $-$ & $-$
& 0.1346 & 65.8139
& 0.1023 & 85.4937
& 0.1744 & 51.5747 \\

ADMMDiff~\cite{admmdiff} & $-$ & $-$
& \underline{0.5586} & \underline{43.9161}
& \underline{0.4932} & \underline{63.2756}
& \underline{0.5755} & \underline{32.2030} \\

\mymodel\ (ours) & $-$ & $-$
& \textbf{0.8032} & \textbf{22.6473}
& \textbf{0.8671} & \textbf{14.9272}
& \textbf{0.7853} & \textbf{20.3174} \\

\bottomrule[1.5pt]
\end{tabular}
}
\end{table}
\begin{table}[t]
\centering
\begin{threeparttable}
\caption{\textbf{Benefits of synthesized images for boosting segmentation performance.} 
(Best result in bold, second-best  underlined. Fully supervised models are listed for reference and not for ranking.)}
\label{tab:dice_aji}
\footnotesize
\setlength{\tabcolsep}{0pt}
\begin{tabular*}{\textwidth}{@{\extracolsep{\fill}} c cc cc cc cc @{}}
\toprule[1.5pt]
\multirow{2}{*}{Method} & \multirow{2}{*}{Training} & \multirow{2}{*}{\shortstack{Paired\\data}}
& \multicolumn{2}{c}{PanNuke}
& \multicolumn{2}{c}{Kumar}
& \multicolumn{2}{c}{MoNuSAC} \\ 
\cmidrule(lr){4-5} \cmidrule(lr){6-7} \cmidrule(lr){8-9}
& & 
& Dice$\uparrow$ & AJI$\uparrow$
& Dice$\uparrow$ & AJI$\uparrow$
& Dice$\uparrow$ & AJI$\uparrow$ \\
\midrule[1pt]

Baseline &  &  &0.7742 &0.5822 &0.8078 &0.5746 &0.7741 &0.5861 \\
CutOut~\cite{cutout}     & $-$ & $-$ &0.8160 &0.6343 &0.8244 &0.6063 &\underline{0.7883} &\underline{0.6046} \\
CutMix~\cite{cutmix}     & $-$ & $-$ &\underline{0.8189} &\underline{0.6368} &\underline{0.8256} &\underline{0.6089} &0.7885 &0.6021 \\

\addlinespace[3pt] 
\hdashline
\addlinespace[3pt]

\textcolor{gray}{SDM~\cite{sdm}}       & \textcolor{gray}{$\circ$} & \textcolor{gray}{$\circ$} & \textcolor{gray}{0.8128} & \textcolor{gray}{0.6375} & \textcolor{gray}{\textbf{0.8280}} & \textcolor{gray}{\textbf{0.6252}} & \textcolor{gray}{\textbf{0.7945}} & \textcolor{gray}{0.6074} \\
\textcolor{gray}{NuDiff~\cite{Nudiff}} & \textcolor{gray}{$\circ$} & \textcolor{gray}{$\circ$} & \textcolor{gray}{0.8223} & \textcolor{gray}{0.6422} & \textcolor{gray}{0.8242} & \textcolor{gray}{0.6097} & \textcolor{gray}{0.7930} & \textcolor{gray}{\textbf{0.6101}} \\

\addlinespace[3pt] 
\hdashline
\addlinespace[3pt]

CycleDiff~\cite{zou2025cyclediff}  & $\circ$ & $-$ &0.8116 &0.6259 &0.8191 &0.6015 &0.7766 &0.5896 \\
SynDiff~\cite{syndiff}     & $\circ$ & $-$ &0.8186 &0.6348 &0.8228 &0.6044 &0.7883 &0.6012 \\
UGDM~\cite{ugdm} & $-$ & $-$         &0.8152  &0.6324  &0.8212  &0.6031  &0.7806  & 0.5884 \\
ADMMDiff~\cite{admmdiff} & $-$ & $-$ &0.8181  &0.6366  &0.8213  &0.6082  &0.7873  &0.6025  \\
\mymodel\ (Ours) & $-$ & $-$ &\textbf{0.8224} &\textbf{0.6437} &\textbf{0.8271} &\textbf{0.6238} &\textbf{0.7895} &\textbf{0.6046} \\
\bottomrule[1.5pt]
\end{tabular*}
\end{threeparttable}
\end{table}
\begin{table}[!ht]
\centering
\begin{threeparttable}
\caption{\textbf{Image quality assessment of synthesized data.}
(Best result in bold, second-best underlined. Fully supervised models are listed for reference and not for ranking.)}
\label{tab:fid_is}
\footnotesize
\setlength{\tabcolsep}{0pt}
\begin{tabular*}{\textwidth}{@{\extracolsep{\fill}} c cc cc cc cc @{}}
\toprule[1.5pt]
\multirow{2}{*}{Method} & \multirow{2}{*}{Training} & \multirow{2}{*}{\shortstack{Paired\\data}}
& \multicolumn{2}{c}{PanNuke}
& \multicolumn{2}{c}{MoNuSAC}
& \multicolumn{2}{c}{Kumar} \\
\cmidrule(lr){4-5}\cmidrule(lr){6-7}\cmidrule(lr){8-9}
& & 

& {\scriptsize FID$\downarrow$} & {\scriptsize IS$\uparrow$}
& {\scriptsize FID$\downarrow$} & {\scriptsize IS$\uparrow$}
& {\scriptsize FID$\downarrow$} & {\scriptsize IS$\uparrow$} \\
\midrule[1pt]

\textcolor{gray}{SDM~\cite{sdm}} & \textcolor{gray}{$\circ$} & \textcolor{gray}{$\circ$}
& \textcolor{gray}{10.0452} & \textcolor{gray}{2.6973}
& \textcolor{gray}{5.7759} & \textcolor{gray}{\textbf{3.2963}}
& \textcolor{gray}{7.8278} & \textcolor{gray}{2.9161} \\
\textcolor{gray}{NuDiff~\cite{Nudiff}} & \textcolor{gray}{$\circ$} & \textcolor{gray}{$\circ$}
& \textcolor{gray}{10.6155} & \textcolor{gray}{3.4152}
& \textcolor{gray}{7.5344} & \textcolor{gray}{2.9655}
& \textcolor{gray}{6.3651} & \textcolor{gray}{2.6072} \\

\addlinespace[3pt] 
\hdashline
\addlinespace[3pt]

CycleDiff~\cite{zou2025cyclediff} & $\circ$ & $-$
& 12.0908 & 2.5262
& 12.2660 & 2.0854
& 12.2660 & 2.3193 \\
SynDiff~\cite{syndiff} & $\circ$ & $-$
& 13.0348 & 1.5896
& 11.1433 & 1.3745
& 11.1433 & 1.7235 \\
UGDM~\cite{ugdm} & $-$ & $-$
& \underline{7.7972} & 3.0428
& 5.9682 & 2.5637
& \underline{5.9761} & 2.3408 \\
ADMMDiff~\cite{admmdiff} & $-$ & $-$
& 9.1775 & \underline{3.1139}
& \underline{5.4475} & \underline{3.1627}
& 6.3033 & \underline{2.8059} \\
\mymodel\ (Ours) & $-$ & $-$
& \textbf{7.7772} & \textbf{4.2514}
& \textbf{5.4099} & \textbf{3.2250}
& \textbf{5.8926} & \textbf{3.4731} \\
\bottomrule[1.5pt]
\end{tabular*}
\end{threeparttable}
\end{table}
\subsection{Alignment Between Prior Mask and Synthesized Image}
\label{Sec:Align}
Inspired by \cite{bhosale2025pathdiff}, we adapt zero-shot nnU-Net \cite{isensee2021nnu} to segment the synthesized images and then evaluate the consistency between the segmentation results and the prior masks. As shown in Table~\ref{tab:fs1_hd95}, our \mymodel\ achieves the best performance in preserving the structural information from the guidance masks. Although GAN-based models excel at style transformation with unpaired data, both SynDiff and CycleDiff fail to regulate mask-to-image correspondence. To refine structural details, UGDM and ADMMDiff steer the sampling trajectory by anchoring the reverse diffusion states. However, their inability to effectively modulate coarse- and fine-grained structures leads to performance degradation. Our \mymodel\ carefully initializes the starting point with structural information and progressively modulates the dynamic states at different frequency components. The performance of our \mymodel\ approaches that of the fully supervised model NuDiff and even exceeds  it on the MoNuSAC dataset. 

\subsection{Improvement on Downstream Segmentation Task}
To demonstrate the benefits of \mymodel\ for downstream segmentation tasks, we augment the training dataset with synthesized images, doubling the size of each dataset. The performance of Hover-Net \cite{graham2019hover} is then compared before and after dataset expansion. In addition to the methods mentioned in Section \ref{Sec:Align}, we also include comparisons with on-the-fly augmentation techniques CutOut~\cite{cutout} and CutMix~\cite{cutmix}. As shown in Table~\ref{tab:dice_aji}, \mymodel\ achieves the highest Dice and AJI scores among all competitive methods. Compared to the baseline, the segmentation performance gain from our framework is substantial, especially considering that it eliminates the need for data annotation and model training. Moreover, we observe that \mymodel's performance approaches that of fully supervised models SDM and NuDiff, which require substantial annotation and computational resources. The performance difference of NuDiff between Table~\ref{tab:fs1_hd95} and Table~\ref{tab:dice_aji} indicates that synthesized images require not only overall contour alignment but also similarity in textural features. Only by achieving both aspects can we avoid introducing distribution shifts during model testing. This finding indirectly validates \mymodel's ability to balance both global structural integration and local feature preservation.

\subsection{Image Quality Assessment of Synthesized Data}
\begin{figure}[t]
    \centering
    \includegraphics[width=\linewidth]{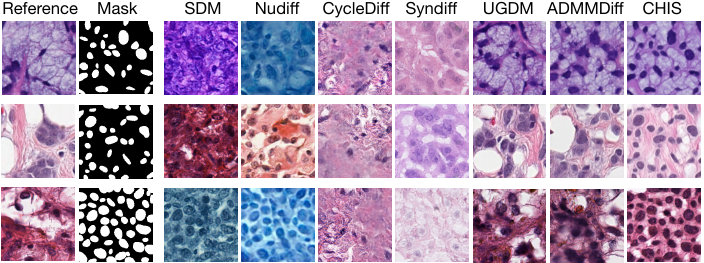}
    \caption{\textbf{Comparison of synthesized images from different methods.}}
    \label{fig:result}
\end{figure}

In this section, we compare the synthesized data directly with the real image dataset, where similarity is quantified using the image quality metrics FID \cite{radford2021learning} and IS \cite{salimans2016improved}. When a style reference image is required for UGDM, ADMMDiff, and \mymodel, we randomly select one from the corresponding training set. Quantitative results in Table~\ref{tab:fid_is} demonstrate the advantage of \mymodel\ in generating high-quality images with respect to visual perception. The improved performance of UGDM, ADMMDiff, and \mymodel\ shows that incorporating a reference during the reverse sampling process effectively regulates the style of generated content. In contrast, both supervised and GAN-based models focus on learning general distribution features from heterogeneous data, which inevitably leads to synthetic data exhibiting artifacts with mixed styles. In Fig.~\ref{fig:result}, we present three example results with their corresponding style references and prior masks. Our \mymodel\ achieves superior performance in terms of both cell distribution and style preservation.

\section{Conclusion}
In this work, we explore the feasibility of a controllable framework for histopathology image synthesis, addressing the critical challenge of limited data resources in clinical applications. Equipped with our training-free structural initialization and textural modulation modules, pretrained diffusion models can now achieve high mask-to-image faithfulness without requiring labor-intensive data annotation or substantial computational overhead. Experimental results demonstrate that our proposed \mymodel\ not only rivals competitive works in generation quality, but also impressively improves downstream instance segmentation performance. As an efficient and scalable solution that bridges the gap between zero-shot efficiency and domain-specific precision, \mymodel\ has the potential to advance various downstream applications, such as virtual staining and pathologist training.

%
%
%

 \bibliographystyle{splncs04}
 \bibliography{CHIS}

\end{document}